\documentclass[letterpaper]{article}
\usepackage{aaai2027}
\usepackage{times}
\usepackage{helvet}
\usepackage{courier}
\usepackage[hyphens]{url}
\usepackage{graphicx}
\usepackage{xcolor}
\usepackage{amsmath,amssymb}
\usepackage{natbib}
\usepackage{caption}
\usepackage{algorithm}
\usepackage{algorithmic}
\usepackage{newfloat}
\usepackage{listings}
\nocopyright  % arXiv 预印版：关闭 AAAI 版权声明脚注
\DeclareCaptionStyle{ruled}{labelfont=normalfont,labelsep=colon,strut=off}
\floatstyle{ruled}
\newfloat{listing}{tb}{lst}{}
\floatname{listing}{Listing}
\title{%Grasp-Guided Last-Mile Navigation for Loco-Manipulation of Legged Robots
LMGN: A Unified Policy Learning Model for Last-Mile Grasp-Navigation
}

\author{
    Xuenmin Hu\textsuperscript{\rm 1},
    Qingxue Qian\textsuperscript{\rm 1},
    Wei Luo\textsuperscript{\rm 1},\\
    Benyi Ma\textsuperscript{\rm 1},
    Bo Tang\textsuperscript{\rm 2},
    Long Chen\textsuperscript{\rm 3}\corresponding
}
\affiliations{
    \textsuperscript{\rm 1}Hubei University, Wuhan, China\\
    \textsuperscript{\rm 2}Worcester Polytechnic Institute, Worcester, MA, USA\\
    \textsuperscript{\rm 3}Institute of Automation, Chinese Academy of Sciences, Beijing, China
}

\begin{document}

\maketitle

\begin{abstract}
Legged robots equipped with manipulators have shown significant potential in the fields of daily life and industry in complex environments. However, existing methods mainly focus on global navigation or manipulation independently, while overlooking the last-mile navigation problem, where the robot must jointly determine the base placement, approach direction, and grasp execution to achieve reliable object grasping, and different manipulation stages require distinct control behaviors. To address these challenges, we propose Last-Mile Grasp-Navigation (LMGN), a unified policy learning model for Quadruped robot loco-manipulation. An Explicit Distance and Directional Reachability (ED-DR) representation is first introduced to explicitly represent the features of grasp-compatible base placement and approach direction. We then develop a semantically aligned Hierarchical Mixture-of-Experts (HMoE) policy that models stage-aware behaviors across the stages of base adjustment, pre-grasp alignment and grasp execution, and object lift. Finally, a new optimization method that jointly aligns navigation and manipulation, stage-dependent expert routing, and action-level consistency is proposed to optimize the HMoE policy. Extensive experiments on the Habitat and Isaac Gym simulators demonstrate that LMGN consistently outperforms state-of-the-art methods, denoting that integrating grasp-aware reachability with stage-aware policy learning significantly improves the success rates of last-mile navigation and target grasp for robots.
\end{abstract}
%\textcolor{red}{The code of LMGN is open-sourced and is available at: \url{https://github.com/QingxueQian/LMGN}.}

\section{Introduction}
Mobile robots that can perceive, move and manipulate are increasingly important for industrial applications, emergency response, space operations, and life services in complex environments~\citep{bellicoso2018advances}. Quadruped robots equipped with manipulators are especially promising because they combine terrain mobility with fine local interaction. They can operate on stairs, slopes, rubble, and narrow passages where conventional wheeled and humanoid robot manipulators are limited, making legged loco-manipulation an active direction in robotics~\citep{bellicoso2019alma, liu2024visual, wang2026odyssey}. %However, a grasp near a table edge illustrates the difficulty. A quadruped robot may observe a bottle near the back of a table, but a small error in base distance or approach direction can place the arm outside a comfortable reach region, direct the gripper toward an obstructed side, or force a after the object has already been disturbed.

Recent studies on quadruped robots equipped with manipulators have made substantial progress in navigation, obstacle traversal, and whole-body control~\citep{wang2025quadwbg}. However, in practical mobile manipulation, task success is not determined only by entering the target neighborhood. The last few steps before grasping are often decisive, which is not only a terminal correction to a path, but also a coupled local decision problem in which the robot must choose an approach direction, assess whether the current stance supports the grasp, decide whether to move the base or the arm first, and recover after a failed grasp. This is the issue of ``last-mile navigation", whose key question is how local motion should be planned to support the final grasp.

Most previous studies related to last-mile navigation focus on local approach~\citep{jauhri2024active}, base placement~\citep{makhal2018reuleaux, shao2026momapos}, pre-grasp positioning~\citep{jauhri2022robot}, and final approach~\citep{zhang2024gamma}. These methods improve local correction near the target, but several limitations remain for legged robots with manipulators. Firstly, navigation and manipulation are often modeled and optimized separately. In this case, a robot may reach the target, but does not stop in a stance that supports the subsequent grasp. Secondly, existing policies often lack explicit models of distance and directional reachability in the last mile. These methods can recognize object shape and pose, but still lack a direct cue for where the robot should stand or which direction should be used for grasping. Thirdly, the loco-manipulation process is strongly stage-dependent, where approach, pre-grasp, grasp, and lift require different policy distributions. Existing loco-manipulation methods often use a single shared network, which may confuse these distributions, leading to the decrease in grasping and navigation success rates.

To address these challenges, we propose Last-Mile Grasp-Navigation (LMGN), a unified policy learning model for robot loco-manipulation. Instead of treating target navigation and grasp execution as separate problems, LMGN integrates an explicit representation for grasp reachability into the last-mile navigation policy and expresses the policy distributions in different manipulation stages with Hierarchical Mixture of Experts (HMoE), so as to make the robot learn to jointly choose poses and approach directions, thus improving the loco-manipulation performance. In conclusion, this paper offers the following three contributions: 1) We propose a unified policy learning model that integrates grasp reachability into the last-mile navigation to improve the loco-manipulation performance. 2) A semantically aligned hierarchical mixture-of-experts policy and a joint optimization method are developed to express the policy distributions in different manipulation stages, improving policy expression. 3) Extensive experiments on the Habitat and Isaac Gym simulators show that LMGN consistently outperforms state-of-the-art methods in terms of gaze and grasp success rates, denoting more reliable and accurate last-mile navigation and object grasp.

\section{Related Work}

\subsection{Legged Loco-Manipulation}
Legged loco-manipulation that combines the terrain mobility of quadrupeds with the local manipulation capability of robotic arms demonstrates considerable potential in industry. Modular methods that decompose the legged loco-manipulation process into separate components including locomotion, perception, planning, and manipulation~\citep{bellicoso2019alma, yokoyama2023asc, wang2025quadwbg} can leverage well-established task-specific controllers and simplify system development, but they suffer from limited coordination between modules and error accumulation. Unified methods study integrated control and long-horizon quadruped manipulation and achieved good coordination of whole-body control by optimizing the policies of leg, torso, and robotic arm~\citep{fu2023deep, yang2023visualforce, liu2024visual}. % but these methods  remain sensitive to high-dimensional control, perception noise, and sim-to-real transfer. %The whole-body methods study integrated control and long-horizon quadruped manipulation~\citep{kuindersma2010control, wang2026odyssey}. 
Compared to modular methods, unified methods effectively improve mobility and coordination, but control policies of different parts of the robots, such as body motion and arm manipulation, are independently optimized with different objectives, leading to the result that the robot can reach the target neighborhood but cannot reach a pose that supports the subsequent grasp. The closest baseline to our method is Visual Whole-Body Control (VBC)~\citep{liu2024visual}, which adopts a hierarchical framework, where a high-level policy predicts the robot base velocity and end-effector pose commands from visual observations, while a low-level whole-body controller tracks these commands for object grasp. However, unlike VBC, our method focuses on last-mile navigation for target grasp, where robot base placement, approach direction, and grasp execution are coupled through reachability-aware features and structured multi-stage policies.

\subsection{Last-Mile Navigation for Robotics}
Last-mile navigation refers to local decision-making after the robot has reached the target neighborhood. Geometry and planning-based methods use reachability maps~\citep{makhal2018reuleaux}, object constraints~\citep{shao2026momapos}, or active perception~\citep{jauhri2024active} to select feasible manipulation poses. These methods are interpretable, but their reliance on explicit scene models, online planning, or task-specific heuristics makes high-frequency closed-loop control difficult in dynamic local scenes. Learning-based methods introduce behavior priors, skill coordination~\citep{jauhri2022robot,hu2023learning,Chen2024Mixed}, or grasp-aware rewards into local navigation~\citep{zhang2024gamma}. Recent manipulation policies further improve robustness through primitive skills~\citep{gu2026diffusion}, flow policies~\citep{zhang2025flowpolicy,Hu2026Enhancing}, one-step learning~\citep{sheng2026mp1}, and self-correction~\citep{pan2025selfcorrecting}. However, most of these methods focus on wheeled or indoor mobile manipulators, fixed-base manipulation, or search for an observable base pose, but do not directly model the coupled choice of quadruped base placement, approach direction, and end-effector pre-grasp motion, resulting in low grasp success rates. We address this gap by converting candidate grasps into navigation-oriented reachability features and proposing a joint optimization framework for navigation and grasp policies combined with a hierarchical MoE to align last-mile navigation with grasp execution.

\section{Problem Formulation}
The last-mile grasp-navigation is a local navigation for the target grasp after the robot enters the target neighborhood. The objective is not the global path search, but the joint choices of base placement, approach direction, and end-effector action within a limited local workspace. These choices should move the robot toward configurations with high grasp reachability. Therefore, we formulate the last-mile grasp-navigation as a cooperative navigation and manipulation problem rather than a local position-correction problem.

Our method is based on reinforcement learning (RL) \cite{hu2026Long}. The last-mile navigation policy aims to select an action to optimize a comprehensive loss $\mathcal{L}_{tot}$ that contains three components: the basic RL loss $\mathcal{L}_{RL}$, the navigation loss $\mathcal{L}_{nav}$, and the grasp loss $\mathcal{L}_{gra}$, as shown by Eq. \ref{eq:last_mile_objective}.

\begin{equation}
\label{eq:last_mile_objective}
\mathcal{L}_{tot}=\lambda_{1}\mathcal{L}_{RL}+\lambda_{2}\mathcal{L}_{nav}+\lambda_{3}\mathcal{L}_{gra},
\end{equation}
where $\lambda_{1}$, $\lambda_{2}$  and $\lambda_{3}$ are the weights for the three losses, respectively.

% maximizes task performance while minimizing the mismatch between the ED-DR feature and the high-level gating distribution, as well as the inconsistency among the commands of the base, end effector, and gripper. Accordingly, the objective is formulated as
% \begin{equation}
% \label{eq:last_mile_objective}
% \begin{aligned}
% a_t^*=\arg\max_{a\in\mathcal{A}}\mathbb{E}\Big[&R_{\rm task}(a\mid s_t)-\lambda_{\rm ali}\Phi_{\rm ali}(a,s_t)\ \\
% &-\lambda_{\rm ac}\Phi_{\rm ac}(a,s_t)\Big],
% \end{aligned}
% \end{equation}
% where $R_{\rm task}$ measures the contribution of the candidate action to task completion. $\Phi_{\rm ali}$ measures the mismatch between the ED-DR feature and the high-level gating distribution. It encourages geometrically similar states to produce similar routing distributions and separates states with distinct recommended base poses. $\Phi_{\rm ac}$ measures the inconsistency in intended actions among the commands of the base, end effector, and gripper, and aligns the three action components with the grasp target. The coefficients $\lambda_{\rm ali}$ and $\lambda_{\rm ac}$ are the weights for $\Phi_{\rm ali}$ and $\Phi_{\rm ac}$, respectively.

\section{Method}

\begin{figure*}[t]
\centering
\includegraphics[width=0.98\textwidth]{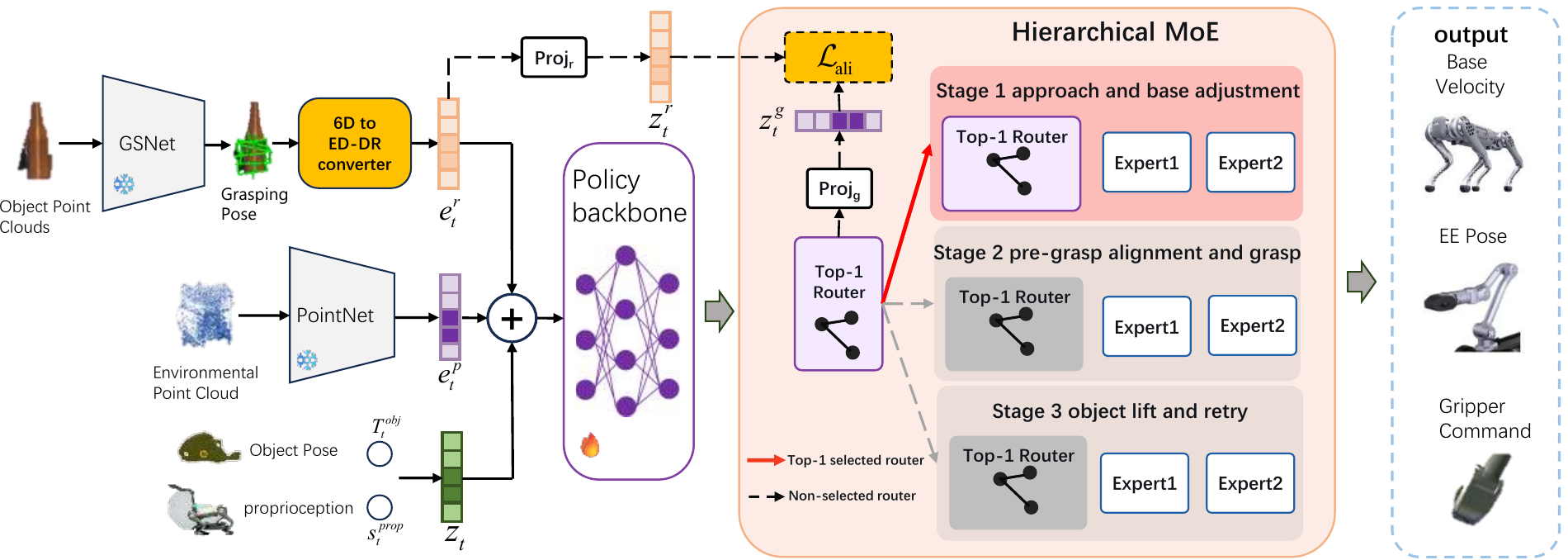}
\caption{Overview of the proposed LMGN. The policy fuses grasp-guided reachability features, point-cloud observations, object pose, and proprioception, then uses a geometry-aligned hierarchical MoE policy to select stage-specific expert routes for base motion, end-effector adjustment, and gripper control.}
\label{fig:gglmn_overview}
\end{figure*}

As shown in Fig.~\ref{fig:gglmn_overview}, the proposed LMGN takes local observations, target pose, and robot proprioceptions as inputs. GSNet~\citep{wang2021graspness} is first used to generate candidate 6D grasp poses from the object point cloud, and then the proposed ED-DR representation converts them into a grasp-guided representation that includes the grasp height, the approach direction, and the recommended relative base placement. Meanwhile, a pretrained PointNet~\citep{qi2017pointnet} is used to extract the surrounding features of the environmental point cloud. Because a feasible grasp also depends on the surrounding geometry and the current robot configuration, the policy backbone that consists of a three-layer multilayer perceptron (MLP) fuses these features with the target pose and robot proprioception. In addition, since the stages of approach, pre-grasp alignment and grasp execution, and the lift or retry require different behaviors, we develop a hierarchical MoE to express policy distributions in different manipulation stages, where semantically aligned Top-1 routing selects a stage-specific expert route and generates the base velocity, end-effector pose increment, and gripper command. In LMGN, navigation--manipulation cooperative optimization integrates stance selection, approach direction adjustment, and end effector grasping in a unified decision objective, so that the policy learns to establish executable grasp rather than merely approach the target.

\subsection{Grasp-Guided Navigation Feature Representation}
\begin{figure}[tb!]
\centering
\includegraphics[width=\columnwidth]{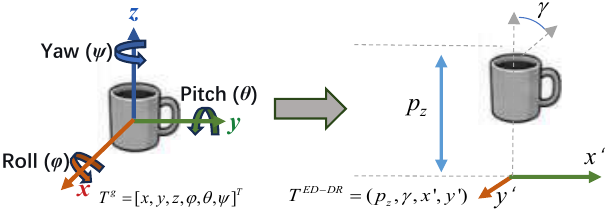}
\caption{ED-DR representation. }%A traditional 6D grasp pose is represented as a compact navigation-oriented feature that contains grasp height, approach direction, and canonical relative base placement.}
\label{fig:ed_dr_rm4d}
\end{figure}

In the last-mile grasp-navigation of robots, the policy needs to infer where the robot should stand, which direction it should approach from, and whether the current base placement allows the grasp to be executed. Existing grasp pipelines often represent a candidate grasp as a 6D pose~\citep{fang2020graspnet, yu2026graphgrasp}, denoted by $T^g=[x, y, z, \phi, \theta, \psi]^{\mathsf T} \in {\rm R}^6$, where $x,y,z$ denote the 3D position coordinates of the end effector, and $\phi,\theta,\psi$ denote the corresponding rotation orientations, as shown in Fig. ~\ref{fig:ed_dr_rm4d}. This representation can express how the end effector should grasp, but cannot represent the relation between base placement and grasp direction, leading to the result that the robot approaches the target without reaching a favorable stance for grasp. 
%In this case, the navigation policy needs to implicitly learn the coupling among the grasp pose, approach direction, and base placement in a high-dimensional space, often leading to the result that the robot approaches the target without reaching a favorable stance for grasping.

Inspired by the low-dimensional pose representation~\citep{rudorfer2025rm4d}, we introduce an Explicit Distance and Directional Reachability (ED-DR) representation in this paper, which converts a conventional 6D grasp pose into geometric features for the last-mile navigation, as shown in Fig. ~\ref{fig:ed_dr_rm4d}. Let $r_z$ denote the grasp approach direction. To obtain a canonical base position, we project the approach direction to the horizontal plane and compute its angle relative to the $x$ axis by Eq. ~\ref{eq:ed_dr_angle}. The relative base position implied by grasping is then rotated into the canonical half-plane aligned with the approach direction, as shown by Eq. ~\ref{eq:canonical_base_rotation}.

\begin{equation}
\gamma={\rm atan2}(r_{z,y},r_{z,x}).
\label{eq:ed_dr_angle}
\end{equation}

\begin{equation}
\label{eq:canonical_base_rotation}
\left[
\begin{array}{c}
x'\\
y'
\end{array}
\right]
=
\left[
\begin{array}{cc}
\cos\gamma & \sin\gamma\\
-\sin\gamma & \cos\gamma
\end{array}
\right]
\left[
\begin{array}{c}
-p_{x}\\
-p_{y}
\end{array}
\right].
\end{equation}

Eqs.~\ref{eq:ed_dr_angle} and~\ref{eq:canonical_base_rotation} convert the candidate 6D grasp pose to last-mile grasp-navigation representation $T^{\rm ED\mbox{-}DR}=(p_{z},\gamma,x',y')$.  $p_{z}$ preserves the target grasp height. $\gamma$ is the approach direction for grasp, and $(x',y')$ denotes the relative base position to execute the grasp. Compared to the conventional 6D pose that is only suitable for grasp, the ED-DR pose, which explicitly contains the appropriate stand position, approach direction and grasp height for grasping the target, is more conducive to loco-manipulation of the last-mile navigation.

%For a candidate grasp pose, the ED-DR feature $z_i^{\rm ED\mbox{-}DR}$ is expressed by $z_i^{\rm ED\mbox{-}DR}=\left(p_{z,i},\theta_i,x_i^*,y_i^*\right)$. %~\ref{eq:ed_dr_feature}.
% \begin{equation}
% \label{eq:ed_dr_feature}
% \begin{array}{rcl}
% z_i^{\rm ED\mbox{-}DR}=\left(p_{z,i},\theta_i,x_i^*,y_i^*\right).
% \end{array}
% \end{equation}
%where $T_i^g \in SE(3)$ is the $i$-th candidate grasp pose in the robot-base or world coordinate frame, and $SE(3)$ is the special Euclidean group of three-dimensional rotations and translations. $p_{z,i}$ is the grasp height, and $\theta_i$ is the angle between the approach direction for grasping and the world vertical axis.

As shown in Fig.~\ref{fig:gglmn_overview}, at time step $t$, the ED-DR representation is encoded as the pose feature $e_t^r$, which is concatenated with the encoded point cloud feature $e_t^p$ and the other feature vector $z_t$ including the target pose $T_t^{\rm obj}$ and the proprioception $s_t^{\rm prop}$, since $T_t^{\rm obj}$ is used to express the orientation of the target object in the local coordinates of the robot, and $s_t^{\rm prop}$ contains the robot base pose, joint positions and velocities, current end-effector pose, and previous action. The concatenated feature is then fed into the policy backbone that follows the method in \citet{liu2024visual} for policy learning.

\subsection{Hierarchical MoE for Multi-Stage Decisions}
To navigate to a precise location and grasp the target, the robot needs to choose an approach direction and adjust the base position, align the end-effector pose and execute grasp, and then lift the object, or recover after a failed grasp. Therefore, a last-mile grasp-navigation often contains the three stages: approach and base adjustment, pre-grasp alignment and grasp execution, and object lift, which have different behavior objectives, ranging from base motion, approach direction to end-effector alignment, and gripper closure. It is difficult for a single shared network to express complex policy distributions, so we need a stage-specific behavior policy.

The Sparse Mixture-of-Experts~\citep{shazeer2017outrageously} routes each input to a small number of specialized experts, allowing different input patterns to be handled separately while limiting interference between inputs and reducing computation consumption. This mechanism provides a natural solution for the multi-stage manipulation in our work, where each stage has a distinct policy distribution and the appropriate stage-specific expert is determined by the current environment state. However, representing different policy distributions between stages and within-stage semantic variations by a single MoE easily entangles the two kinds of factors, weakening the performance of MoE representation.

%Sparse Mixture-of-Experts~\citep{fedus2022switch}, which conditionally activates specialized experts for different tasks, reducing interference among stage-dependent behaviors while maintaining efficient routing. Different manipulation stages in this conditional routing mechanism correspond to distinct policy distributions that benefit from dedicated expert representations, which provides a natural solution: the sparse MoE routes heterogeneous inputs to dedicated experts, while multi-stage manipulation requires state-dependent selection among distinct policy modes. However, representing different policy distributions between stages and semantics variations in a stage by a flat MoE easily entangle the two kinds of factors, weakening the performance of MoE representation. 

%Sparse Mixture-of-Experts~\citep{fedus2022switch}, which conditionally activates specialized experts for different states, reducing interference among stage-dependent behaviors while maintaining efficient routing. This conditional routing mechanism, where different manipulation stages correspond to distinct policy distributions that benefit from dedicated expert representations, provides a natural solution.  \textcolor{red}{The key similarity lies in conditional specialization: sparse MoE routes heterogeneous inputs to dedicated experts, while multi-stage manipulation requires state-dependent selection among distinct policy modes.} 

To address this issue, we propose a hierarchical MoE (HMoE) for multi-stage loco-manipulation. As shown in Fig.~\ref{fig:gglmn_overview}, the high-level experts express the three stage-special tasks including approach and base adjustment, pre-grasp alignment and grasp execution, and object lift, as well as automatically identify different manipulation stages through the semantically aligned features, while the low-level experts represent within-stage semantic variations for action execution. This design models decision-making for last-mile grasp-navigation as a stage-conditioned semantic control problem. Given the fused feature $z_t$, the hierarchical MoE is expressed by Eq.~\ref{eq:moe_decision}.

%\{color{red}It conditionally activates specialized experts for different states, reducing interference among stage-dependent behaviors while maintaining efficient Top-1 routing. This also avoids averaging expert outputs with incompatible physical semantics, making sparse MoE suitable for multimodal last-mile loco-manipulation.}
% {\color{red}However, a flat MoE represents stage semantics and within-stage geometric variations in the same routing space, which may entangle the two factors and weaken expert specialization. Meanwhile, assigning only one expert to each stage cannot capture different geometric action modes within that stage. We therefore introduce a hierarchical MoE, where the high-level experts model stage-dependent behaviors and the low-level experts capture
% within-stage geometric variations.}
%By factorizing stage selection and within-stage geometric specialization, the hierarchical structure produces stage-consistent yet geometry-adaptive decisions.

\begin{equation}
\label{eq:moe_decision}
\begin{array}{rcl}
k_t&=&{\rm Top1}(g_h(z_t)), \quad  m_t={\rm Top1}(g_l(z_t|k_t)),  \\
a_t&=&f_{HMoE}(z_t|k_t,m_t),
\end{array}
\end{equation}
where $k_t$ and $m_t$ are the indexes of the selected high-level stage expert and the low-level semantic sub-expert at time step $t$, respectively. $g_h$ and $g_l$ are the expert route selection functions of the high- and low-level MoEs, respectively. $f_{HMoE}$ denotes the policy network of the hierarchical MoE. ${\rm Top1}(\cdot)$ denotes activating the expert with the highest gating probability. To avoid semantic mixing caused by averaging the outputs of multiple experts, only one expert route generates the action $a_t=(v_t^{\rm base},\Delta T_t^{\rm ee},c_t^{\rm grip})$, where $v_t^{\rm base}$, $\Delta T_t^{\rm ee}$, and $c_t^{\rm grip}$ denote the base velocity, the end-effector pose increment, and the gripper command, respectively. The Top1 MoE in our method follows the expert-routing structure in~\citep{ma2026cmoecontrastivemixtureexperts}. 
%whose structure follows the expert-routing design in CMoE~\citep{ma2026cmoecontrastivemixtureexperts}. 

% The MoE model uses Top-1 sparse routing rather than weighted averaging because action semantics differ across stages and gripper commands are partly discrete. Top-1 routing keeps base motion, end-effector adjustment, and gripper action on the same expert path.

\subsection{Joint Optimization for Grasp-Navigation}
HMoE can provide an effective decision ability for expressing stage-independent tasks and within-stage semantic variations in terms of network architecture, but relying solely on the architecture cannot ensure that expert routing selection follows stage-independent tasks or within-stage semantic relations since the HMoE policy still needs to be optimized by effective objectives. Therefore, we propose an effective joint optimization method for grasp-navigation, including the following three components.

\subsubsection{RL optimization for last-mile grasp-navigation: }
Following the baseline VBC~\citep{liu2024visual}, we train the policy backbone with Proximal Policy Optimization (PPO)~\citep{schulman2017proximal}, where the reward function is a significant factor for policy learning. However, the reward in VBC only provides delayed feedback for basic manipulation tasks without navigation feedback. Therefore, we design an RL optimization method for last-mile grasp-navigation. The reward function is shown as Eq. \ref{eq:reward}.

%and does not distinguish an arbitrary nearby stance from a grasp-friendly one, we augment it with navigation-quality and grasp-reachability rewards:
\begin{equation}
\label{eq:reward}
r_t=r_t^{\rm task}+\lambda_{\rm nav}r_t^{\rm nav}
+\lambda_{\rm gra}r_t^{\rm gra},
\end{equation}
where $r_t^{\rm task}$ is the task reward in VBC, which provides feedback on grasp, lift, and the success or failure of the final task. $r_t^{\rm nav}$ and $r_t^{\rm gra}$ are the designed navigation and grasp rewards, respectively. $\lambda_{\rm nav}$ and $\lambda_{\rm gra}$ are their weights and fixedly set to $1.0$ and $0.5$, respectively, to balance the reward values. $r_t^{\rm nav}$ and $r_t^{\rm gra}$ are calculated by Eqs. \ref{eq:navigation_reward} and \ref{eq:grasp_reward}, respectively.

\begin{equation}
\label{eq:navigation_reward}
\begin{array}{rcl}
r_t^{\rm nav}
=\left(
\left\|\mathbf{b}_{t-1}-\mathbf{b}_t^*\right\|_2
-\left\|\mathbf{b}_t-\mathbf{b}_t^*\right\|_2
\right)\\
+\beta_{\rm y}\left(
d_{\rm y}(\gamma_{t-1},\gamma_t^*)
-d_{\rm y}(\gamma_t,\gamma_t^*)
\right), 
\end{array}
\end{equation}

% \begin{equation}
% \label{eq:navigation_reward}
% \begin{array}{rcl}
% r_t^{\rm nav}
% =\left(
% \left\|\mathbf{b}_{t-1}-\mathbf{b}_t^*\right\|_2
% -\left\|\mathbf{b}_t-\mathbf{b}_t^*\right\|_2
% \right)\\
% +\beta_{\rm y}\left(
% d_{\rm y}(\gamma_{t-1},\gamma_t^*)
% -d_{\rm y}(\gamma_t,\gamma_t^*)
% \right), \\
% d_{\rm y}(\gamma_1,\gamma_2)=|\operatorname{atan2}(\sin(\gamma_1-\gamma_2),\cos(\gamma_1-\gamma_2))|,
% \end{array}
% \end{equation}

\begin{equation}
\label{eq:grasp_reward}
r_t^{\rm gra}
=\mathcal{R}_{\rm ED\mbox{-}DR}\left(B_t,T_t^{g*}\right)
-\mathcal{R}_{\rm ED\mbox{-}DR}\left(B_{t-1},T_t^{g*}\right).
\end{equation}
where $\mathbf{b}_t=(x'_t,y'_t)$ denotes the planar position of the base in ED-DR representation at time step $t$, and $\mathbf{b}_t^*$ is its corresponding target position. $B_t=(\gamma_t,x'_t,y'_t)$ denotes the pose of the base in ED-DR representation at time step $t$, and $B_t^*$ is its corresponding target pose for the selected grasp pose $T_t^{g*}$ in 6D pose representation. $\gamma_t,\gamma_t^*\in(-\pi,\pi]$ are the yaw angles. Positions and yaw angles are measured in meters and radians, respectively, in the ED-DR coordinates. The function $d_{\rm y}(\cdot)$ computes the shortest wrapped angular distance. $\beta_{\rm y}$ is a weight and fixedly set to 0.2 to balance positional and yaw-alignment progress. The normalized score $\mathcal{R}_{\rm ED\mbox{-}DR}(B,T^{g*})\in[0,1]$, as calculated by Eq. \ref{eq:R_ED-DR}, measures the grasp reachability of a candidate grasp $T^{g*}$ from a base pose $B$, with a higher score indicating greater reachability. %The 2nd term in Eq. \ref{eq:grasp_reward} is calculated 

\begin{equation}
\label{eq:R_ED-DR}
\mathcal{R}_{\rm ED\mbox{-}DR}(B,T^{g*})
=\exp\!\left[-\left(\left\|\mathbf{b}-\mathbf{b}^*\right\|_2 \\
+\beta_{\rm y}d_{\rm y}\!\left(\gamma,\gamma^*\right)\right)\right].
\end{equation}

\subsubsection{Alignment optimization between high-level expert routes and ED-DR feature: }
HMoE provides separate expert routes matching with different stages for loco-manipulation policies, where the high-level expert routes express stage-dependent behaviors of the robots. Therefore, how to couple the high-level expert routing with the manipulation stage is an important issue. Since the loco-manipulation stage is decided by the states and observations of the robot, it is necessary to align observations with high-level expert routes in HMoE to enhance the matching ability between the loco-manipulation stages and the expert routes, thus improving the policy performance in different stages.

In this paper, we design an alignment optimization between high-level expert routes and ED-DR feature. As shown in Fig.~\ref{fig:gglmn_overview}, at time step $t$, the ED-DR feature $e_t^r$ and high-level gating distribution $e_t^g$ are projected into a shared space by the two projection heads in multi-positive contrastive learning (MPCL)~\citep{khosla2020supervised}, generating $z_t^{\rm r}$ and $z_t^{\rm g}$, respectively. For the $i$-th sample in a minibatch with $\mathcal{M}$ samples, $\mathbf{b}_i^*$ and $\gamma_i^*$ are the planar position and yaw angle of the target base pose, respectively. We define $\mathcal{P}(i)$ as the positive set which contains the $i$-th sample and all samples from the same stage to the $i$-th sample, and $\mathcal{N}(i)$ as the negative set which contains all samples from different stages with the $i$-th sample. Each sample in the minibatch constructs a positive set and a negative set, so we can obtain $\mathcal{M}$ positive sets and negative sets,  based on which the alignment loss $\mathcal{L}_{\rm ali}$ is defined by Eq. \ref{eq:geo_loss}.
\begin{equation}
\label{eq:geo_loss}
\begin{aligned}
\mathcal{L}_{\rm ali}
=&-\frac{1}{|\mathcal{M}|}
\sum_{i\in\mathcal{M}}
\frac{1}{|\mathcal{P}(i)|}
\sum_{p\in\mathcal{P}(i)}
\log\\[-1mm]
&\frac{
\exp\!\left(\operatorname{sim}(z_i^{\rm r},z_p^{\rm g})/\tau\right)
}{
\displaystyle\sum_{a\in\mathcal{P}(i)\cup\mathcal{N}(i)}
\exp\!\left(\operatorname{sim}(z_i^{\rm r},z_a^{\rm g})/\tau\right)
},
\end{aligned}
\end{equation}
where $\operatorname{sim}(\cdot)$ denotes cosine similarity, and $\tau$ is the temperature parameter set to $0.2$ in our method. This objective in Eq. $\mathcal{L}_{\rm ali}$ is used to train the high-level expert routes in HMoE by the self-supervised learning mechanism in MPCL~\citep{khosla2020supervised}, aligning observations in different stages with the corresponding expert routes in the high-level HMoE, thus learning different policy distributions for different stages.

\subsubsection{Action-level consistency optimization: }
The proposed RL optimization method considers both navigation and grasp actions, but the rewards of navigation and grasp are calculated separately and then integrated into the total reward function, which cannot guarantee that the commands of the three components, including the base, end effector (arm), and gripper, are compatible and remain consistent to carry out a certain action in each stage. For example, the base is moving toward a position and has not reached the appropriate grasp position, but the end effector and the gripper are attempting to grasp the target. Such inconsistency in the intended actions for commands of different components often results in failures of the last-mile grasp-navigation task, even when each individual command appears reasonable.

To ensure that the commands of the three components are consistent in the intended actions, we develop an action-level consistency optimization method. We define $\Delta\mathbf{b}_t=\mathbf{b}_t-\mathbf{b}_t^*$ at time step $t$, denoting the increment from the base position $\mathbf{b}_t$ to the target base position $\mathbf{b}_t^*$ in the ED-DR coordinates. Let $T_t^{\rm g}$ be the grasp pose in the 6D pose representation at time step $t$. $\Delta T_t^{\rm g*}$ is the increment of the grasp pose from $T_t^{\rm g}$ to the target grasp pose $T_t^{g*}$. 

The loss function in the baseline method VBC considers three terms including the base position increment, the gripper pose increment, and the opening and closing state of the gripper. In order to employ different learning objectives in different stages and make the commands of the three components remain consistent in the intended actions, we weight different terms in the original loss function by gating probability of the selected high-level expert routes for stage-special policy learning. Therefore, the proposed action-level consistency loss $\mathcal{L}_{ac}$ is defined by Eq. \ref{eq:consistency_loss}.
\begin{equation}
\label{eq:consistency_loss}
\begin{aligned}
\mathcal{L}_{ac}={}&w_t^{\rm g}\alpha_b\Bigl(1-
\operatorname{cos}\bigl(\Delta\mathbf{b}_t,\Delta\mathbf{b}_t^*\bigr)\Bigr)\\
&+w_t^{\rm g}\alpha_g\!\left\|\Delta T_t^{\rm g}\!-\!\Delta T_t^{\rm g*}\right\|_2^2
+\alpha_c\,\mathcal{L}_{\rm BCE}\bigl(c_t^{\rm g},c_t^*\bigr),
\end{aligned}
\end{equation}
where $c_t^{\rm g}$ is the predicted gripper-closing probability. $c_t^* \in \{0, 1\}$ is a binary value, denoting the stage-dependent gripper target state. The weight $w_t^{\rm g}$ is the gating probability from the high-level expert routes. The function $\operatorname{cos}(\cdot)$ is the cosine similarity, and $\mathcal{L}_{\rm BCE}$ is the binary cross-entropy loss.  $\alpha_b$, $\alpha_g$, and $\alpha_c$ are the weights for the three terms and set to $1.0$, $0.2$, and $0.2$, respectively, in all experiments following the settings in VBC.

\subsubsection{Joint Optimization Objective}
In conclusion, the total joint optimization objective $\mathcal{L}_{\rm tot}$ in Eq. \ref{eq:last_mile_objective} is evolved into a specific implementation, as shown by Eq. \ref{eq:training_objective}.
\begin{equation}
\label{eq:training_objective}
\mathcal{L}_{\rm tot}
=\lambda_{\rm RL}\mathcal{L}_{\rm RL}
+\lambda_{\rm ali}\mathcal{L}_{\rm ali}
+\lambda_{ac}\mathcal{L}_{ac},
\end{equation}
where $\mathcal{L}_{\rm RL}$ is the standard PPO loss computed using the reward in Eq. ~\ref{eq:reward}. $\lambda_{\rm RL}$, $\lambda_{\rm ali}$, and $\lambda_{ac}$ are the weights for $\mathcal{L}_{\rm RL}$, $\mathcal{L}_{\rm ali}$, and $\mathcal{L}_{ac}$, respectively, and are uniformly set to $1.0$, $0.1$, and $0.2$ in all experiments without fine-tuning.

%The PPO objective provides the primary task-level learning signal, while the two auxiliary losses impose semantic structure on the routing space and coordinate the base, end-effector, and gripper actions around the same selected grasp target. The policy network, expert routes, and routing modules are jointly optimized under Eq.~\ref{eq:training_objective}.

\section{Experiments}
\subsection{Experimental Setup}
We evaluated the method on the Habitat~\citep{szot2022habitat20traininghome} and Isaac Gym~\citep{makoviychuk2021isaac} simulators both using a Unitree B1 robot with a Z1 arm, which has a total of 19 controllable DOF including 12 DOF for the four legs, 6 DOF for the arm, and 1 DOF for the gripper. For every episode, the base pose, target pose, and local obstacle layout are randomly initialed, and the robot is required to avoid obstacles and grasp the target on tabletops. If the robot successfully grasps and stably lifts the target within the specified number of steps, the task is determined to be successful. If the target drops, the step number exceeds the maximum, or the end effector cannot form a valid grasp pose, the task is determined to be failed. Details of experimental configurations and hyper-parameters are provided in Appendixes A and C, respectively.

Following the methods VBC~\citep{liu2024visual} and GAMMA~\citep{zhang2024gamma}, two metrics including GraspSR and GazeSR are used in this paper. GraspSR is the ratio of successful grasp number to the total trial number, measuring whether the robot completes gripper closure, target grasping, and stable lifting. GazeSR is the ratio of successful gaze number to the total trial number, measuring whether the robot reaches a precise state for grasp. A successful gazing is defined that the wrist camera is within 15~cm of the target object and its viewing-direction error is smaller than $10^\circ$ when the grasp command is triggered following the settings in GAMMA.

SOTA methods including VBC~\citep{liu2024visual}, HB~\citep{szot2022habitat20traininghome}, ReachMM~\citep{jauhri2022robot}, and GAMMA~\citep{zhang2024gamma} are used as the comparative methods. VBC is a legged loco-manipulation method and also uses a hierarchical policy architecture. HB is a standard loco-manipulation baseline that learns a monolithic policy to jointly control the mobile base and arm. ReachMM is a reachability-aware loco-manipulation method that leverages the target position and arm reachability priors to guide base placement and arm reaching. GAMMA is also one of the most classic graspability-aware loco-manipulation method. %We report the performance of these methods using their best results from their original papers. VBC did not report the results in this benchmark, so we report its results under the same configurations as our method.

Following the baselines, the policy is trained for 60,000 steps using 10,240 parallel environments. Each episode lasts at most 150 environment steps unless the task succeeds or fails. All experiments are conducted on an NVIDIA GeForce RTX 4090 GPU with 24~GB memory and a part of Intel Xeon Gold 6530 CPU with 30 cores. %Details of software and hardware equipment are provided in Appendix C.

\subsection{Comparative Results with SOTA Methods}
For fair comparisons with the above mentioned SOTA methods, we conducted our experiments on Habitat under exactly the same simulator setup, robot models, and protocols, as the settings used in GAMMA. Table~\ref{tab:gamma_comparison} shows the comparative results with the SOTA methods. 

LMGN outperforms all comparative methods, achieving a GazeSR of 73.9\% and a GraspSR of 74.8\%. HB and ReachMM perform worse than the other three methods in both metrics, since the generation of candidate grasp poses is lacking in the two methods, which seriously affects the accuracy of subsequent grasp execution. Compared to the baseline VBC, LMGN improves GazeSR and GraspSR by 4.9 and 22 percentage points, respectively. Compared to GAMMA, LMGN improves GazeSR and GraspSR by 7.6 and 10.3 percentage points, respectively. The improvement in GazeSR suggests that ED-DR provides clearer guidance for selecting base positions and approach directions that facilitate grasp. The larger improvement in GraspSR further suggests that the proposed HMoE effectively models stage-special policies, while joint optimization coordinates base motion, end-effector adjustment, and gripper control toward the same grasp target. Consequently, LMGN enables the robot not only to reach a local state suitable for grasp but also to reliably translate that state into a successful grasp in complex loco-manipulation tasks.

\begin{table}[tb!]
\centering
\begin{tabular}{lcc}
\hline
Method & GazeSR (\%) $\uparrow$ & GraspSR (\%) $\uparrow$ \\
\hline
VBC & 69.0 & 52.2 \\
HB & 31.9 & -- \\
ReachMM & 21.5 & 12.7 \\
GAMMA & 66.3 & 64.5 \\
LMGN (ours) & \textbf{73.9} & \textbf{74.8} \\
\hline
\end{tabular}
\caption{Comparative results with SOTA methods on Habitat. ``-" indicates no reported data in the original paper.}
\label{tab:gamma_comparison}
\end{table}

\subsection{Ablation Studies}
\paragraph{Ablation studies of main modules:}
To evaluate the contributions of the main modules of our method, we compare LMGN with its four variants, including 1) the baseline VBC, 2) LMGN without the ED-DR representation (\textit{w/o ED-DR}), 3) LMGN where HMoE is replaced by a traditional MLP (\textit{w/o HMoE}), and 4) LMGN without joint optimization (\textit{w/o JO}). In ablation studies, we follow the baseline VBC and conduct experiments in another simulator, Isaac Gym, which is more challenging than Habitat because all the joint torques in the robots need to be controlled through a low-level whole-body controller rather than directly controlling the base velocity, the end-effector pose, and the gripper command. Moreover, we evaluate six kinds of different objects that cover regular, elongated, asymmetric, hollow, and slender geometries, and separately report the GazeSR and GraspSR metrics of these objects to examine whether each module remains effective under different base placement, approach direction, and grasp requirements. Tables ~\ref{tab:gazesr_ablation} and ~\ref{tab:graspsr_ablation} show the results of GazeSR and GraspSR, respectively. 

Tables~\ref{tab:gazesr_ablation} and~\ref{tab:graspsr_ablation} show that LMGN achieves the highest average GazeSR and GraspSR of 84.47\% and 53.86\%, whereas \textit{w/o JO} performs worst, with those of 26.67\% and 2.79\%, respectively, because removing joint optimization seriously disrupts coordination among the base, end effector, and gripper. \textit{w/o HMoE} achieves a higher GazeSR on Bowl (49.53\% versus 38.08\% of LMGN) because the rotationally symmetric geometry permits multiple feasible approach directions that could not satisfy the strict GazeSR criterion, though LMGN achieves a higher GraspSR. \textit{w/o HMoE} also achieves a slightly higher GraspSR on Cup (60.90\% versus 57.83\% of LMGN), since the simple geometry requires less stage specialization.

\begin{table}[!t]
\centering
\setlength{\tabcolsep}{3pt}
\begin{tabular}{lrrrrr}
\hline
Object & VBC & \shortstack{w/o\\HMoE} & \shortstack{w/o\\ED-DR} & \shortstack{w/o\\JO} & \shortstack{LMGN\\(Ours)} \\
\hline
Square Box & 69.04 & 45.22 & 39.81 & 32.83 & \textbf{100.00} \\
Long Box & 31.16 & 7.75 & 12.39 & 9.49 & \textbf{100.00} \\
Drill & 30.84 & 9.30 & 17.95 & 28.02 & \textbf{76.92} \\
Cup & 61.95 & 60.26 & 41.99 & 39.22 & \textbf{97.89} \\
Bowl & 34.79 & \textbf{49.53} & 30.62 & 29.90 & 38.08 \\
Bottle & 38.28 & 13.44 & 17.80 & 20.54 & \textbf{93.90} \\ \hline
Mean & 44.34 & 30.92 & 26.76 & 26.67 & \textbf{84.47} \\
%Std & 15.30 & 21.30 & 11.41 & 9.49 & 22.21 \\
\hline
\end{tabular}
\caption{GazeSR (\%) results of ablation study for different objects on Isaac Gym.}
\label{tab:gazesr_ablation}
\end{table}

\begin{table}[!t]
\centering
\setlength{\tabcolsep}{3pt}
\begin{tabular}{lrrrrr}
\hline
Object & VBC & \shortstack{w/o\\HMoE} & \shortstack{w/o\\ED-DR} & \shortstack{w/o\\JO} & \shortstack{LMGN\\(Ours)} \\
\hline
Square Box & 23.05 & 56.25 & 39.20 & 2.77 & \textbf{57.42} \\
Long Box & 16.26 & 41.61 & 30.11 & 0.20 & \textbf{46.01} \\
Drill & 15.56 & 38.08 & 26.25 & 0.97 & \textbf{46.38} \\
Cup & 21.12 & \textbf{60.90} & 41.32 & 3.83 & 57.83 \\
Bowl & 23.23 & 60.37 & 46.71 & 7.81 & \textbf{62.85} \\
Bottle & 19.81 & 43.18 & 28.42 & 1.13 & \textbf{52.68} \\ \hline
Mean & 19.84 & 50.07 & 35.34 & 2.79 & \textbf{53.86} \\
%Std & 3.02 & 9.35 & 7.50 & 2.55 & 6.17 \\
\hline
\end{tabular}
\caption{GraspSR (\%) results of ablation study for different objects on Isaac Gym.}
\label{tab:graspsr_ablation}
\end{table}

\paragraph{Component-wise ablation in joint optimization:}
To further evaluate the effect of the three components in the joint optimization for grasp-navigation, we conduct a component-wise ablation study where keeping the ED-DR representation and HMoE architecture unchanged. Specifically, we compare the full joint optimization with three ablated variants, including 1) the variant that removes the additional rewards for navigation and grasp reachability in RL optimization (\textit{w/o Reward}), 2) the variant that removes the alignment optimization objective \textit{w/o RouteAlign}, and 3) the variant that removes the action-level consistency optimization objective \textit{w/o ActionCons}. All variants use the same training configuration, random seeds, environment settings, and evaluation protocol as the full model.

Tables~\ref{tab:conm_component_gazesr} and~\ref{tab:conm_component_ablation} show that the full joint optimization method achieves the highest average GazeSR and GraspSR of 84.47\% and 53.86\%, respectively. Removing any component substantially reduces both metrics, with \textit{w/o ActionCons} causing the largest decreases to 19.10\% and 6.59\% for the average GazeSR and GraspSR, respectively, which verifies the importance of coordinating the base, end effector, and gripper. The full joint optimization method also achieves the highest GraspSR for all objects and the highest GazeSR for five objects. The bowl is an exception, as its multiple feasible approach directions weaken the correspondence between the GazeSR metric and actual grasp success.

\begin{table}[!t]
\centering
\setlength{\tabcolsep}{3pt}
\small
\begin{tabular}{lrrrr}
\hline
Object & \shortstack{w/o\\ActionCons} & \shortstack{w/o\\Reward} & \shortstack{w/o\\RouteAlign} & \shortstack{Joint \\optimization} \\
\hline
Square Box & 19.12 & 31.40 & 46.76 & \textbf{100.00} \\
Long Box & 14.15 & 16.50 & 32.63 & \textbf{100.00} \\
Drill & 11.17 & 10.40 & 25.81 & \textbf{76.92} \\
Cup & 23.00 & 34.47 & 51.03 & \textbf{97.89} \\
Bowl & 34.28 & 47.21 & \textbf{51.68} & 38.08 \\
Bottle & 12.90 & 18.34 & 32.21 & \textbf{93.90} \\ \hline
Mean & 19.10 & 26.39 & 40.02 & \textbf{84.47} \\
%Std & 7.87 & 12.53 & 10.17 & 22.21 \\
\hline
\end{tabular}
\caption{GazeSR (\%) of component-wise ablation methods in joint optimization for different objects.}
\label{tab:conm_component_gazesr}
\end{table}

\begin{table}[!t]
\centering
\small
\setlength{\tabcolsep}{3pt}
\begin{tabular}{lrrrr}
\hline
Object & \shortstack{w/o\\ActionCons.} & \shortstack{w/o\\Reward} & \shortstack{w/o\\RouteAlign} & \shortstack{Joint \\optimization} \\
\hline
Square Box & 3.56 & 4.51 & 32.77 & \textbf{57.42} \\
Long Box & 0.05 & 0.15 & 1.68 & \textbf{46.01} \\
Drill & 0.23 & 0.43 & 3.50 & \textbf{46.38} \\
Cup & 12.80 & 16.24 & 34.09 & \textbf{57.83} \\
Bowl & 22.60 & 24.56 & 45.41 & \textbf{62.85} \\
Bottle & 0.30 & 0.38 & 9.74 & \textbf{52.68} \\ \hline
Mean & 6.59 & 7.71 & 21.20 & \textbf{53.86} \\
%Std & 8.43 & 9.41 & 16.89 & 6.17 \\
\hline
\end{tabular}
\caption{GraspSR (\%) of component-wise ablation methods in joint optimization for different objects.}
\label{tab:conm_component_ablation}
\end{table}

\subsection{Generalization Experiments for Unseen Objects}
To evaluate the generalization of LMGN beyond the objects used for training, we test the trained policy without fine-tuning on three unseen objects including clear box, plastic banana, and pitcher. We use the same evaluation protocol and report GazeSR and GraspSR for each object in Table~\ref{tab:unseen_objects}. LMGN achieves average values of 77.49\% and 48.52\% for GazeSR and GraspSR, respectively, on unseen objects, which are close to 84.47\% and 53.86\% that are the results on the training objects, demonstrating that the proposed LMGN has a good generalization.

\begin{table}[!t]
\centering
\setlength{\tabcolsep}{3pt}
\begin{tabular}{lrr}
\hline
Object & GazeSR & GraspSR \\
\hline
Clear Box & 82.40 & 41.90 \\
Plastic Banana & 75.05 & 46.82 \\
Pitcher & 75.02 & 56.83 \\
\hline
Mean & 77.49 &48.52 \\
\hline
\end{tabular}
\caption{Success rates (\%) for unseen objects.}
\label{tab:unseen_objects}
\end{table}

%The consistently high GazeSR indicates effective generalization in approach and base-placement decisions. The lower GraspSR on the grey plate and Phillips screwdriver suggests that precise grasp execution is more sensitive to unseen thin or slender geometries.
% grey plate & 70.38 & 15.04 \\
% phillips screwdriver & 68.76 & 28.68 \\\
% Mean & 74.32 & 37.85 \\

\subsection{Expert Behavior Visualization in Different Stages}
\begin{figure}[tb!]
\centering
\includegraphics[width=1.0\columnwidth]{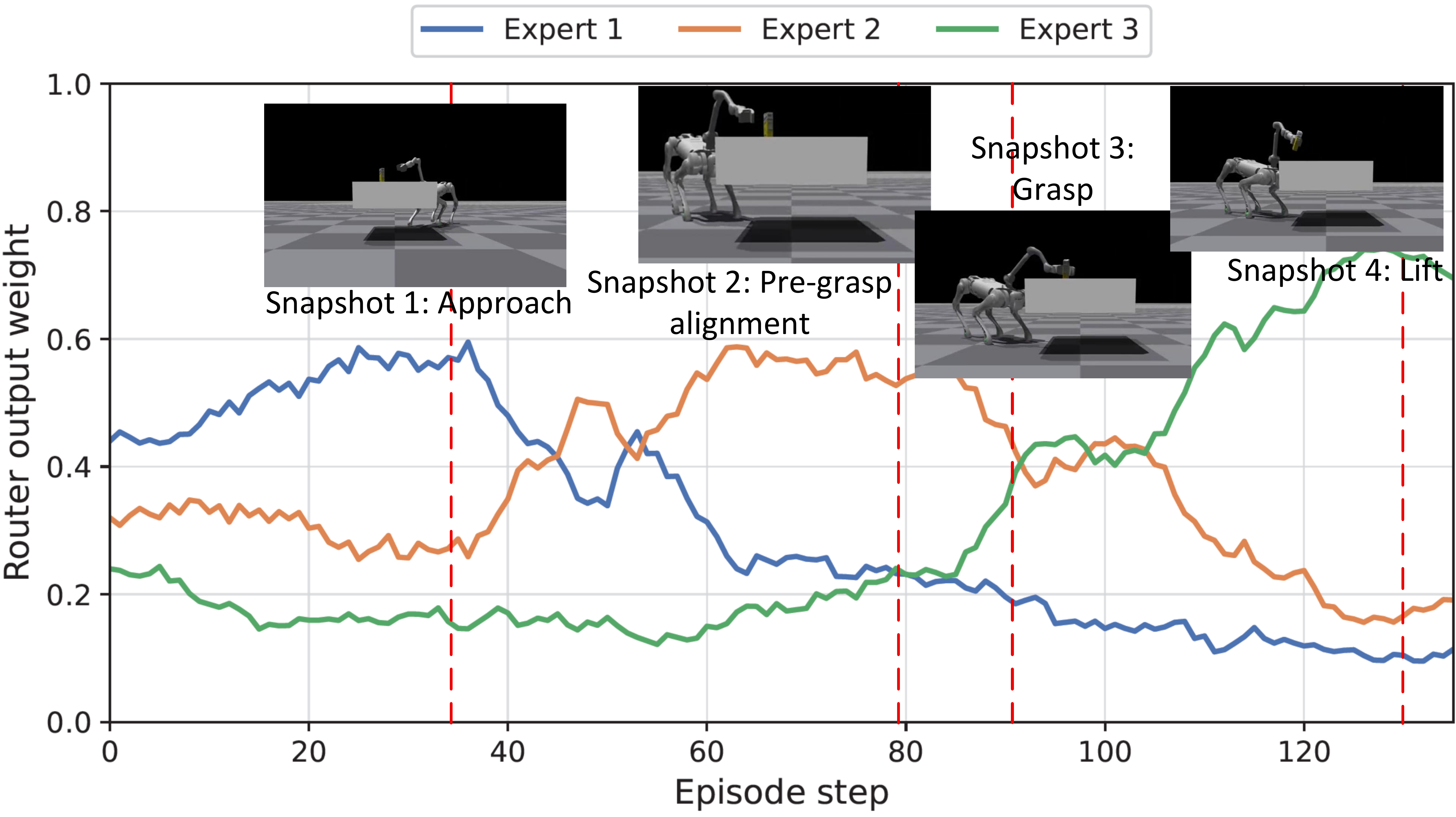}
\caption{Weight curves of the high-level expert routes in HMoE and some snapshots during a representative episode.}
\label{fig:expert_behaviorandSnap_analysis}
\end{figure}

To evaluate the effect of the proposed HMoE, we visualize the weight changes of the high-level expert routes in HMoE at different stages. Figure~\ref{fig:expert_behaviorandSnap_analysis} shows the weight curves and some snapshots during a representative episode. Expert~1 has the highest weight in the early steps, corresponding to the approach and base adjustment stage (Snapshot 1). Expert~2 becomes dominant in the intermediate steps, corresponding to the pre-grasp alignment and grasp execution stage (Snapshots 2 and 3). Expert~3 gradually becomes dominant after approximately step 105, corresponding to the lift stage (Snapshots 4). Although there is some overlap at the transitions between different stages, the changes in the expert weights are consistent with the order of the loco-manipulation stages, which shows that the high-level expert routes match different stages and HMoE can express different policy distributions in different stages.

%The policy first moves the robot toward a grasp-compatible base placement, then refines the end-effector pose for pre-grasp alignment, and finally closes the gripper and lifts the object. In particular, the intermediate alignment stage separates base adjustment from gripper closure and demonstrates coordinated base, arm, and gripper behaviors. This staged progression is consistent with the stage-dependent control intended by the hierarchical MoE policy.

% \begin{figure}[tb!]
% \centering
% \begin{tabular}{@{}cc@{}}
% \includegraphics[width=0.42\columnwidth,height=0.27\columnwidth,keepaspectratio]{figures/stage_base_approach.png} &
% \includegraphics[width=0.42\columnwidth,height=0.27\columnwidth,keepaspectratio]{figures/stage_end_effector_alignment.png} \\
% {\small (a) Approach (Stage 1)} & {\small (b) Pre-grasp alignment (Stage 2)} \\[4pt]
% \includegraphics[width=0.42\columnwidth,height=0.27\columnwidth,keepaspectratio]{figures/stage_grasp.png} &
% \includegraphics[width=0.42\columnwidth,height=0.27\columnwidth,keepaspectratio]{figures/stage_lift.png} \\
% {\small (c) Grasp (Stage 2)} & {\small (d) Lift (Stage 3)}
% \end{tabular}
% \caption{Snapshots of the last-mile navigation for loco-manipulation in different stages. %(a) Base approach. (b) End-effector alignment. (c) Grasp. (d) Lift.
% }
% \label{fig:qualitative_episode}
% \end{figure}

\section{Conclusions}
In this paper, we propose LMGN, a unified policy framework for last-mile navigation and manipulation of legged robots. An ED-DR representation that transforms candidate 6D grasp poses into navigation-oriented geometric cues is first introduced to explicitly express the features of grasp-compatible base placement and approach direction. We then develop a semantically aligned Hierarchical MoE policy that models stage-specific behaviors across loco-manipulation stages including base adjustment, pre-grasp alignment and grasp execution, and object lift. To optimize the HMoE policy, we also propose a joint optimization method that contains RL optimization for grasp-navigation, alignment optimization between expert routes and ED-DR features, and action-level consistency. Experimental results on Habitat and Isaac Gym show that LMGN outperforms  SOTA methods in metrics of GazeSR and GraspSR, and ablation studies confirm the contributions of the unified policy learning model. %Future work will study real-robot deployment under perception errors and dynamic scenes.

\bibliography{aaai2026}

\end{document}